\documentclass[sigconf]{acmart}

\AtBeginDocument{%
  }

\setcopyright{acmlicensed}
\copyrightyear{2018}
\acmYear{2018}
\acmDOI{XXXXXXX.XXXXXXX}

\acmConference[Conference acronym 'XX]{Make sure to enter the correct
  conference title from your rights confirmation emai}{June 03--05,
  2018}{Woodstock, NY}
\acmISBN{978-1-4503-XXXX-X/18/06}

\usepackage[export]{adjustbox}
\input{preamble.tex}
\usepackage{array}
\newcolumntype{P}[1]{>{\centering\arraybackslash}p{#1}}

\newtheorem*{assumption*}{\assumptionnumber}

\newcommand{\bert}{BERT\xspace}
\definecolor{arsenic}{rgb}{0.23, 0.27, 0.29}

\definecolor{silver}{rgb}{0.75, 0.75, 0.75}
\newcommand{\wpred}[1]{
\textcolor{silver}{#1}\xspace
}
\begin{document}
\title{\algtitle}

%%
%% The "author" command and its associated commands are used to define
%% the authors and their affiliations.
%% Of note is the shared affiliation of the first two authors, and the
%% "authornote" and "authornotemark" commands
%% used to denote shared contribution to the research.
\author{Anshul Mittal}
\email{anshulmittal@microsoft.com}
\author{Shikhar Mohan}
\author{Deepak Saini}
\author{Siddarth Asokan}
\affiliation{%
  \institution{Microsoft}
  \country{India}
}

\author{Lakshya Kumar}
\author{Pankaj Malhotra}
\author{Jian Jiao}
\author{Amit Singh}
\affiliation{%
  \institution{Microsoft}
  \country{India}
}

\author{Suchith Chidananda Prabhu}
\author{Sumeet Agarwal}
\affiliation{%
  \institution{IIT Delhi}
  \country{India}
}

\author{Soumen Chakrabarti}
\affiliation{%
  \institution{IIT Bombay}
  \country{India}
}

\author{Purushottam Kar}
\affiliation{%
  \institution{IIT Kanpur}
  \country{India}
}

\author{Manik Varma}
\email{manik@microsoft.com}
\affiliation{%
  \institution{Microsoft, IIT Delhi}
  \country{India}
}

% \author{Valerie B\'eranger}
% \affiliation{%
%   \institution{Inria Paris-Rocquencourt}
%   \city{Rocquencourt}
%   \country{France}
% }

% \author{Aparna Patel}
% \affiliation{%
%  \institution{Rajiv Gandhi University}
%  \city{Doimukh}
%  \state{Arunachal Pradesh}
%  \country{India}}

% \author{Huifen Chan}
% \affiliation{%
%   \institution{Tsinghua University}
%   \city{Haidian Qu}
%   \state{Beijing Shi}
%   \country{China}}

% \author{Charles Palmer}
% \affiliation{%
%   \institution{Palmer Research Laboratories}
%   \city{San Antonio}
%   \state{Texas}
%   \country{USA}}
% \email{cpalmer@prl.com}

% \author{John Smith}
% \affiliation{%
%   \institution{The Th{\o}rv{\"a}ld Group}
%   \city{Hekla}
%   \country{Iceland}}
% \email{jsmith@affiliation.org}

% \author{Julius P. Kumquat}
% \affiliation{%
%   \institution{The Kumquat Consortium}
%   \city{New York}
%   \country{USA}}
% \email{jpkumquat@consortium.net}

%%
%% By default, the full list of authors will be used in the page
%% headers. Often, this list is too long, and will overlap
%% other information printed in the page headers. This command allows
%% the author to define a more concise list
%% of authors' names for this purpose.
\renewcommand{\shortauthors}{Mittal et al.}

%%
%% The abstract is a short summary of the work to be presented in the
%% article.

%%
%% The code below is generated by the tool at http://dl.acm.org/ccs.cfm.
%% Please copy and paste the code instead of the example below.
%%


%%
%% Keywords. The author(s) should pick words that accurately describe
%% the work being presented. Separate the keywords with commas.
\keywords{Extreme classification, Large scale recommendation, Metadata, Sponsored search, ads, intelligent advertisement}
%% A "teaser" image appears between the author and affiliation
%% information and the body of the document, and typically spans the
%% page.

%%
%% This command processes the author and affiliation and title
%% information and builds the first part of the formatted document.

% \maketitle

% \renewcommand{\shortauthors}{A. Mittal et. al.}
\begin{abstract}
Extreme Classification (XC) offers a scalable and efficient solution for retrieving highly relevant ads in Sponsored Search settings, significantly enhancing user engagement and ad performance. Most tasks in sponsored search involve highly skewed distributions over the data point (query) and label (ads) space with limited or no labeled training data. One approach to tackle this long-tail classification problem is to use additional data, often in the form of a graph such as similar queries, same session queries \textit{etc.} that are associated with user queries/ads, called graph metadata. Graph-based approaches, particularly Graph Convolutional Networks (GCNs), have been successfully proposed to leverage this graph metadata and improve classification performance. However, for tail inputs/labels, GCNs induce graph connections that can be noisy, leading to downstream inaccuracies while also incurring significant computation and memory overheads. To address these limitations, we introduce a novel approach, RAMEN, that harnesses graph metadata as a regularizer while training a lightweight encoder rather than a compute- and memory- intensive GCN-based method. This avoids the inaccuracies incurred by noisy graph induction and sidesteps the computational costs of GCNs via an easy-to-train and deploy encoder. The proposed approach is a scalable and efficient solution that significantly outperforms GCN-based methods. Extensive A/B tests conducted on live multi-lingual Bing Ads search engine traffic revealed that RAMEN increases revenue by \textbf{1.25-1.5\%} and click-through rates by \textbf{0.5-0.6\%} while improving quality of predictions across different markets. Additionally, evaluations on public benchmarks show that RAMEN achieves up to \textbf{5\%} higher accuracy compared to state-of-the-art methods while being \textbf{50\%} faster to infer, and having \textbf{70\%} fewer parameters. 
\end{abstract}
\maketitle

\section{Introduction}
\label{sec:Intro}

Sponsored search is a key revenue driver for search engines, displaying ads alongside organic results.  The task in Sponsored Search is that of understanding the user query and recommending relevant ads. One of the popular approaches to predicting these recommendations is extreme classification. Extreme classification (XC) refers to a supervised machine learning paradigm wherein multi-label learning is performed on extremely large label spaces. The ability of XC to handle enormous label sets with millions of labels makes it an attractive choice for applications such as product recommendation~\citep{Medini19,Dahiya21,Mittal22,Kharbanda2022}, search \& advertisement~\citep{Prabhu18b,Dahiya21,Jain16}, and query recommendation~\citep{Jain19,Chang20}. The key appeal of XC comes because of two reasons: (a) The ability to recommend relevant ads to user queries even if queries are previously unseen (tail queries), and (b) The ability to accurately tag rare/tail ads relevant to a user query. An ad/query is considered to be part of the tail if very few training data points are associated with it. The tail problem is further aggravated due to the issue of \emph{missing data}, since tail query/ads are also at a higher risk of going missing~\citep{Jain16} in the ground truth. In solving the tail-data problem, XC approaches rely on metadata. Beyond textual ads descriptions~\citep{Mittal21,Dahiya21b,Dahiya23}, this auxiliary metadata can augment the limited supervision available for tail query/ads and is typically available in the form of multi-modal descriptions such as images~\citep{Mittal22}, or graphs~\citep{Mittal21b,Saini21, Mohan2024}. In this paper, we focus on graph metadata which is available in several applications, \textit{e.g.} for online search and recommendation, users asking the multiple queries in the same search session (co-session queries) can be a metadata graph over queries.

\begin{figure}
    \centering
      \includegraphics[width=0.8\linewidth]{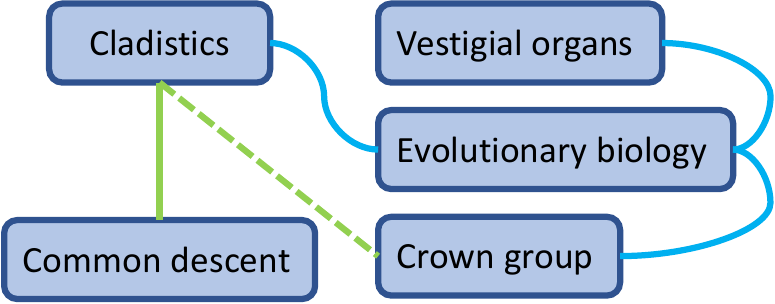}
      \includegraphics[width=0.7\linewidth]{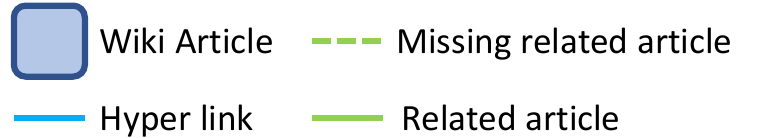}
    \caption{A snapshot from LF-WikiSeeAlsoTitles-320K dataset for the article on ``Cladistics.'' The related article ``Common descent'' is tagged but the ground truth is missing the link to ``Crown group''. Traversal on the hyperlink edges can help discover missing link but can also lead to irrelevant link such as ``Vestigial organs''.}
    \label{fig:datasnapshot}
\end{figure}

\noindent \textbf{Graph metadata in XC:} Graph metadata has been used in XC to (a) handle tail ads~\citep{Mikel21, Mittal21b, Kharbanda2022}, and (b) enhance the user query representations~\citep{Mohan2024, Saini21, yang2021graphformers, Chien2023}. To handle tail ads, these approaches use textual descriptions of an ad along with graph metadata to learn ad embeddings via graph convolutional networks (GCN). To enhance user query representation, these algorithms rely on a two stage retrieval pipeline wherein, for an unseen query, first the graph metadata nodes are retrieved, and subsequently, a GCN combines them with the query representation. The new combined representation is then used in a second stage to retrieve the relevant set of ads. The retrieval of graph nodes or graph traversal can also help discover missing ads associated with query. To showcase this, let us consider an example from a publicly available dataset from XC~\citep{XMLRepo} called LF-WikiSeeAlsoTitles-320K. Here, task is similar to that of query-to-ads recommendation \textit{i.e.} for a Wikipedia document title retrieve related Wikipedia page. For training, the ground truth comprises links to multiple Wikipedia page under the \textit{``See Also''} section of the Wikipedia page. This ground truth is often incomplete and contains lots of missing links. Each Wikipedia page also has multiple hyperlinks to other Wikipedia page, which can be used to construct the metadata graph. A snapshot of the dataset is provided in \figurename~\ref{fig:datasnapshot}, and shows how a rare Wikipedia page (which was previously unseen in the training ground truth) ``Crown group'' can be recovered for the Wikipedia article ``Cladistics'' by traversing the aforementioned graph. Similar to hyperlink graphs, in sponsored search, for queries/ads, we have graphs of (a) slight textual perturbations (similar queries/ads) but having similar search intent; (b) queries that were asked in the same session, or (c) ads that were clicked in the same search session. \alg utilize these metadata graphs to make faster and accurate predictions in comparison to GCNs like OAK~\citep{Mohan2024}.

\begin{figure}
\centering
  \includegraphics[width=0.95\linewidth]{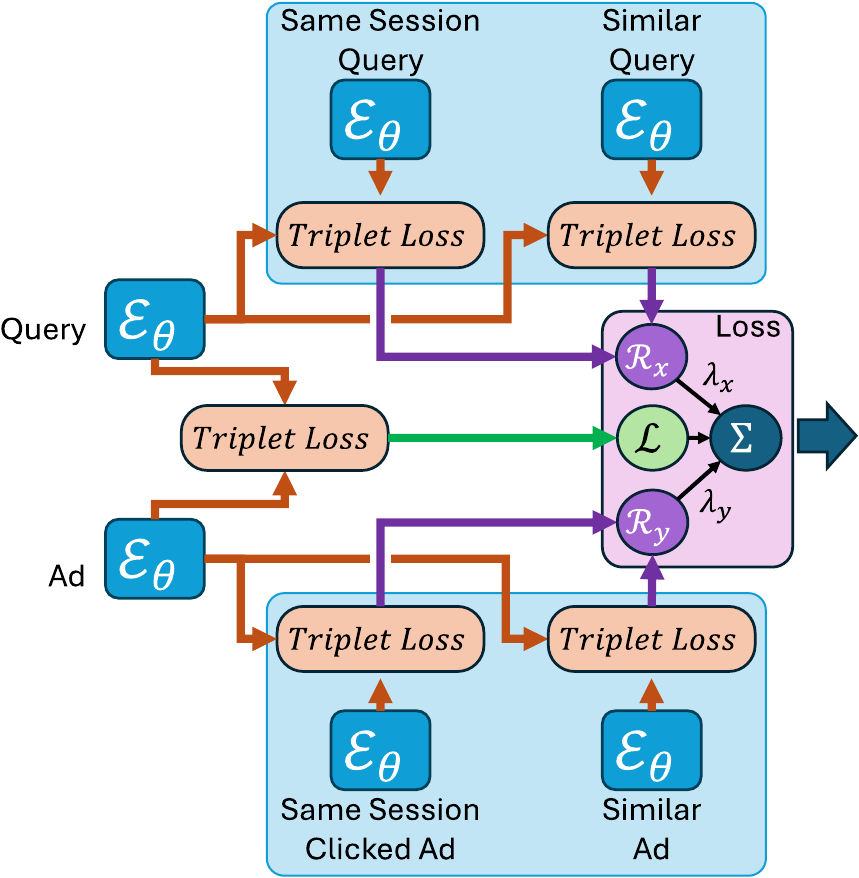}
\caption{ \alg uses graph metadata to regularize encoder during training. Training framework of \alg is robust to noise in graph. Here $\cR_x$ and $\cR_y$ are regularization loss terms and $\cL$ is the task based loss term.
}
\label{fig:training}  
\end{figure}

 \begin{figure*}
\centering
  \includegraphics[width=0.85\linewidth]{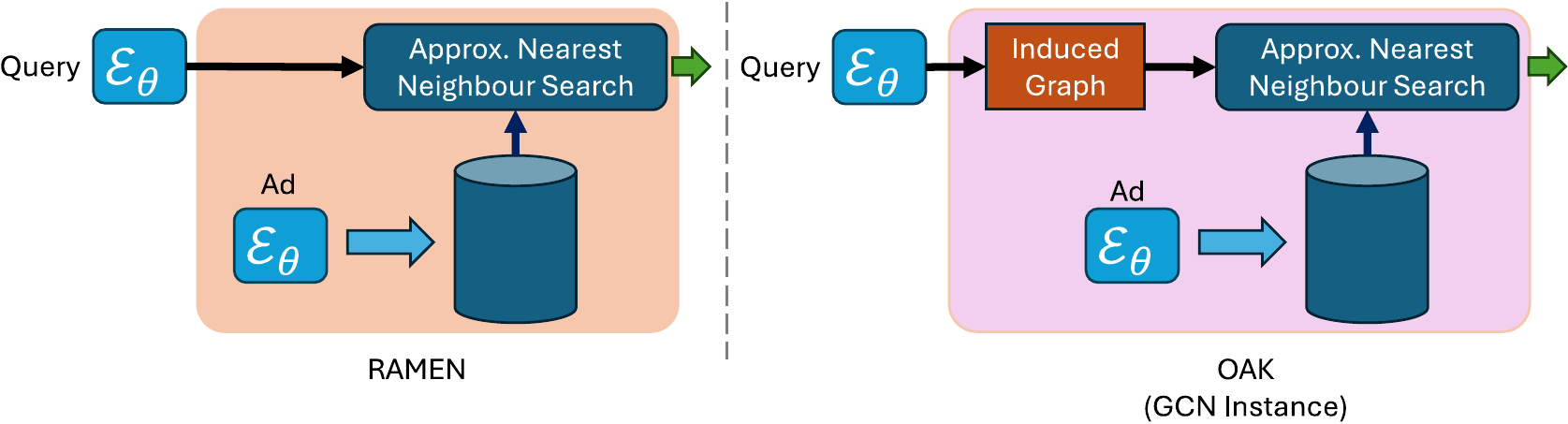}
  \caption{  \alg does not require additional information to compute an accurate representation of the test point. In OAK (an instance of a GCN), inference is a computationally expensive two-stage pipeline, where the test point is first embedded into the graph and then, the linked nodes are used to compute the final representation (induced graph). Since \alg uses the graph for regularization, and does not traverse the graph at inference time, it can be 2$\times$ faster, and 5\% more accurate, than OAK.
   }
  \label{fig:compare}  
\end{figure*}

\noindent\textbf{Limitations of GCN Methods}: In online scenarios, new users and new advertisers are continuously using the Bing search engine. The **new user queries and ads account for 75-80\% for the search volume. Having graph metadata associated with each of the items is a challenging tasks. GCNs circumvent this challenge by inducing graphs for both queries and ads. These induced graphs are noisy and can lead to wrong predictions. To illustrate this, recall the LF-WikiSeeAlsoTitles-320K hyperlink graph -- Traversal over the graph can also lead to irrelevant links such as ``Vestigial organs'' and extracting meaningful information from such noisy graphs is a challenge. It has been noted that having a high-quality graph can offer enhanced model accuracy in recommendation settings~\citep{Mittal21b,Saini21,yang2021graphformers, Chien2023}. However, such quality is rarely available for a new query or ad. In addition to this, the use of GCN architectures makes both training and inference 50\% and 100\% more expensive (Table~\ref{tab:train_pred_time}) as compared to \alg which is a dual encoder model. Noisy graphs, along with expensive inference architecture, make GCNs inaccurate for tail and costly to deploy. Our primary research question is: \textbf{\textit{How can we use graph metadata for accurate predictions on tail queries and ads while saving computational cost compared to GCNs?}}

% a method of biological classification based on common ancestry, r
\subsection{Our Contributions} 
To address the above question, we propose g\textbf{R}aph regul\textbf{A}rized encoder training for extre\textbf{ME} classificatio\textbf{N} (\textbf{\alg}). \alg is a framework to effectively utilize graph metadata during training, with minimal overheads while saving computational cost in model size and inference time (Table~\ref{tab:train_pred_time}) as compared to GCNs. In particular \alg's contributions are as follows:
\begin{itemize}
    \item \textbf{Graph-based regularization}: Using graph metadata to regularize the training of a dual encoder (Figure~\ref{fig:training}). This improves the model's ability to capture semantic relationships between queries/ads, even for tail queries/ads which resulted in 1.25-1.5\% revenue gains in over 160+ countries and up to 4.8\% reduction in brand mismatch rates (Table~\ref{fig:bing_bmr}).
    \item \textbf{Robustness to graph noise}: Developing a mechanism to dynamically adjust the reliance on graph metadata during training based on the confidence in the graph links (Table~\ref{tab:ablation}). This helps mitigate the impact of noisy graphs (Table~\ref{tab:oracleLinker}) in training model parameters.
    \item \textbf{Efficient inference}: \alg provides accurate retrievals for tail queries and avoids the costly step of inducing graph as used in GCNs like OAK (Figure~\ref{fig:compare}). This reduces computation cost at inference time by 50\%. (Table~\ref{tab:train_pred_time}).
\end{itemize}
\alg scales to datasets with up to 360M ads and can offer up to 5\% higher prediction accuracies over state-of-the-art methods including those that use graph metadata to train GCN.

% To address the above question, we propose g\textbf{R}aph regul\textbf{A}rized encoder training for extre\textbf{ME} classificatio\textbf{N} (\textbf{\alg}). \alg is a framework to effectively utilize graph metadata at scale, with minimal overheads in training cost and zero overhead in model size or inference time (Table~\ref{tab:train_pred_time}). \alg can be incorporated into any existing XC system in a modular manner with few alterations (Table~\ref{tab:results-benchmark}). \alg can handle multiple graphs -- graphs over **users/data points, graphs over ads/labels, or both -- and offers superior prediction accuracy, even when presented with noisy graphs (Figure~\ref{fig:training} \& Section~\ref{sec:exp}). While the \alg encoder is trained using the metadata graph, during inference, unlike **baseline GCNs, the \alg framework does **invovle graph traversal, and therefore does not require graphs. This reduces \alg inference to inference on a dual encoder, significantly improving latency (Figure~\ref{fig:compare}). \alg scales to datasets with up to 360M labels and can offer up to **15\% higher prediction accuracies over state-of-the-art methods including those that use graph metadata to train GCN.  Code for \alg will be released publicly. 

\section{Related work}
\label{sec:related}

\textbf{Extreme classification (XC)} is a key paradigm in several areas such as ranking and recommendation. The literature on XC methods is vast~\citep{Medini19,Dahiya21,Babbar17,You18,Prabhu18b,Jain16,Jain19,Guo19,Mittal21,Mittal21b, Saini21, Wydmuch18, Liu17, Jiang21,Chalkidis19,Ye20,Zhang21, Mineiro15,Jasinska16,Khandagale19,Yen17,Wei19,Siblini18a, Gupta19, Gupta2023}. Early XC methods used fixed (bag-of-words) ~\citep{Babbar17,Prabhu18b,Jain16,Khandagale19,Yen17,Wei19} or pre-trained~\citep{Jain19} features and focused on learning only a classifier architecture. Recent advances have demonstrated significant gains by using task-specific features obtained from a variety of deep encoders such as bag-of-embeddings~\citep{Dahiya21,Dahiya23}, CNNs~\citep{Liu17}, LSTMs~\citep{You18}, and transformers~\citep{Jiang21,Chalkidis19,Ye20,Zhang21}. Training is scaled to millions of labels and training points by performing encoder pre-training followed by classifier training~\citep{Dahiya21}. A data point is trained only on its relevant labels (that are usually few in number) and a select few irrelevant labels deemed most informative (known as \textit{hard negatives}) and obtained using a process known as \emph{negative mining} ~\citep{Mikolov13,Guo19,Xiong20,Dahiya21b,Dahiya23, Faghri18,Chen20,He20,Karpukhin20,Lee19,Luan20,hofstatter21,Qu21}.

\noindent \textbf{Label Metadata in XC}: Most XC methods use textual representations as label metadata since they facilitate scalable training and inference and allow for leveraging good-quality pre-trained deep encoders such as RoBERTa~\citep{Liu19roberta}, DistilBERT base~\citep{Sanh2019DistilBERTAD}, etc. Examples include encoder-only models such as DEXML~\citep{gupta24}, TwinBERT~\citep{Lu20} and ANCE~\citep{Xiong20}, and encoder+classifier architectures such as DECAF~\citep{Mittal21}, SiameseXML~\citep{Dahiya21b}, X-Transformer~\citep{Chang19}, XR-Transformer~\citep{Chang20}, LightXML~\citep{Jiang21}, and ELIAS~\citep{Zhang21}, amongst many others~\citep{Ye20,Liu19,You18,Chalkidis19}. There is far fewer works in the literature on the use of other forms of label metadata. For instance, ECLARE~\citep{Mittal21b} and GalaxC~\citep{Saini21} use graph convolutional networks whereas MUFIN~\citep{Mittal22} explores multi-modal label metadata in the form of textual and visual descriptors for labels. 

% For use of label text in recommendation, contributions from NLR literature~\citep{Devlin19,He21,Liu19roberta,Yang19,Clark20,Fedus21,meng2021cocolm,Lu20,Liu21,Lee21} must also be noted. However, Much effort has gone into making NLR models larger with the goal of reaching human parity on various natural language understanding and inference tasks~\citep{Wang19}. However, scaling these large NLR models can be a challenging task. 

\noindent \textbf{Graph Neural Networks in Related Areas}: A sizeable body of work exists on using graph neural networks such as graph convolutional networks (GCN) for recommendation~\citep{yang2021graphformers, Mohan2024, graphsage, fastgcn, ladies, asgcn, clustergcn, graphsaint, zhu2021textgnn, he2020lightgcn, yang2022kgcl}. Certain methods, \textit{e.g.,} FastGCN~\citep{fastgcn}, KGCL~\citep{yang2022kgcl}, and LightGCN~\citep{he2020lightgcn} learn item embeddings as (functions of) free vectors. This makes them unsuitable for making predictions on a novel test point. Other GCN-based methods such as OAK~\citep{Mohan2024}, PINA~\citep{Chien2023}, GraphSAGE~\citep{graphsage} and GraphFormers~\citep{yang2021graphformers} learn node representations as functions of node metadata e.g. textual descriptions. Consequently, these methods work in zero-shot settings, but they still incur the high storage and computational cost of GCNs. Moreover, diminishing returns are observed with increasing the number of layers of the GCN~\citep{Mittal21b,clustergcn} with at least one model, namely LightGCN~\citep{he2020lightgcn} foregoing all non-linearities in its network, effectively opting for a single-layer GCN. It must be noted that GCNs can be highly accurate if one has access to an oracle for predicting relevant nodes (Table~\ref{tab:oracleLinker}). However, such oracle is never available online and the slightest error in first stage retrieval leads to poor retrieval quality (Sec.~\ref{sec:exp}). 

We now develop the \alg method that offers a far more scalable alternative to GCNs and other popular graph-based architectures in XC settings, while significantly reducing the overheads of graph-based learning, and offering sustained and significant performance boosts in prediction accuracies.

\section{\alg: g\underline{R}aph regul\underline{A}rized encoder training for extre\underline{ME} classificatio\underline{N}}
\label{sec:alg}

\textbf{Notation:} Let $L$ be the number of labels in the recommendation task. Let $\vx_i, \vz_l$ be the textual descriptions of the queries or data point $i$ and ads or label $l$ respectively. From this point, we will be using query as data point and ads as labels interchangeably as per context need. For each data point $i \in [N]$, its ground-truth label vector is $\vy_i \in\{-1,+1\}^L$, where $y_{il} = +1$ if label $l$ is relevant to the data point $i$ and otherwise $y_{il} = -1$. The training set is comprised of $N$ labeled data points and $L$ labels as $\cD:=\{\{\vx_i,\vy_i\}_{i=1}^{N},\{{\vz_l}\}_{l=1}^{L}\}$. Let $\cX \deff \{\vx_i\}_{i=1}^{N}$ denote the set of training data points and $\cZ \deff \{\vz_l\}_{l=1}^{L}$ denote the set of labels. The metadata graph over the anchor sets $\cA$ (**hyper-links, co-bidded queries) is denoted by $\cG_{XA}$ and $\cG_{ZA}$ for data point (users) and label (ads), respectively. 

\noindent \textbf{Metadata Graphs}: In recommendation scenario, \alg obtains metadata graphs over queries and ads using user session or textual similarity as described in introduction section. These graph is essentially links between queries/ads to their relevant node in the graph. These are also called \emph{Anchor Sets}. Refer to \figurename~\ref{fig:datasnapshot}, here hyper-linked Wikipedia pages are called anchor set. Let $\cA = \bc{\va_1,\va_2,\ldots,\va_M}$ denote an anchor set of $M$ elements \textit{e.g.} pages connected via hyperlink for LF-WikiSeeAlsoTitles-320K dataset. We abuse notation to let $\va_m$ denote the textual representation of anchor item $m \in [M]$ as well. Each query and ad is associated with metadata graph over anchor sets:
\begin{enumerate}[leftmargin=*]
    \item Query metadata graph: This is denoted as $\cG_{XA} = (V_{XA},E_{XA})$ with $V_{XA} \deff \cX \cup \cA$ i.e., the union of training data points and anchor points. The matrix $E_{XA} = \bc{e_{im}} \in \bc{0,1}^{N\times M}$ encodes whether data point $\vx_i$ has an edge to anchor item $\va_m$ or not.
    \item Ads metadata graph: This is denoted as $\cG_{ZA} = (V_{ZA},E_{ZA})$ with $V_{ZA} \deff \cZ \cup \cA$ i.e. the union of labels and anchor points. The matrix $E_{ZA} = \bc{e_{lm}} \in \bc{0,1}^{L\times M}$ encodes whether label $\vz_l$ has an edge to anchor item $\va_m$ or not.
\end{enumerate}

\subsection{Regularized Training Framework}
\alg incorporates graph based regularization while training model parameters. \alg training requires two components: (a) A base XC ($\cM$) component which consists of an encoder block ($\cE_{\vtheta}$), and (b) The metadata graph ($\cA_c$) component as described above. The encoder $\cE_{\vtheta}: \cX \rightarrow \cS^{D-1}$ with trainable parameters $\vtheta$ is used to embed query and ads using their textual descriptions. Here, $\cS^{D-1}$ denotes the $D$-dimensional unit sphere, \emph{i.e.,} the encoder provides unit-norm embeddings (unless stated otherwise). \alg uses a DistilBERT~\citep{Sanh2019DistilBERTAD} encoder as $\cE_{\vtheta}$. In the following sections, we first explain training framework of \alg followed by incorporating graph regularization for robustness.

\noindent \textbf{\alg loss function}: \figurename~\ref{fig:training} shows overall training framework for \alg. The encoder is trained using triplet loss $(\cL(\vtheta))$ over query and ad representation, regularized using two components: a) The anchor set on the query side $(\cR_x(\vtheta))$, and b) The anchor set on ad side $(\cR_z(\vtheta))$, as explained in metadata graph section. The $\cL(\vtheta)$ function is then given by:
\begin{equation*}
    \label{eq:siamese}
    \cL(\vtheta) = 
 \sum_{i = 1}^N\sum_{\substack{l: y_{il} = +1\\k: y_{ik} = -1}} [\cE_{\vtheta}(\vz_k)^\top\cE_{\vtheta}(\vx_i) - \cE_{\vtheta}(\vz_l)^\top\cE_{\vtheta}(\vx_i) + \gamma]_+.
\end{equation*}
Note that this loss function encourages the encoder to embed a query close to its relevant ad and far from irrelevant ones. The optimized encoder ($\vtheta^{*})$ through \alg framework is obtained by minimizing the following objective
\[
\vtheta^{*} = \min_{\vtheta} \Big\{\cL(\vtheta) + \sum_{t=1}^{T} \left(\lambda_{x}\cR^{t}_{x}(\vtheta) + \lambda_{z} \cR^{t}_z(\vtheta)\right)\Big\},
\]
where $\lambda_{x}$ and $\lambda_{z}$ are regularization constants. $\cR^{t}_x(\vtheta)$ and $\cR^{t}_z(\vtheta)$ are regularization loss applied over anchor set $\cA_t$ as explained below.

\noindent\textbf{Metadata Graph Regularizers}: Given an encoder \(\cE_{\vtheta}\), an anchor set $\cA$, and graphs $\vG_{XA}, \vG_{ZA}$, we define the following two regularization functions over the encoder parameters:
\begin{align*}
    \cR_x(\vtheta) &= \textstyle
 \sum\limits_{i = 1}^N\sum\limits_{\substack{p: e_{ip} = 1\\n: e_{in} = 0}} [\cE_{\vtheta}(\vx_i)^\top\cE_{\vtheta}(\va_n) - \cE_{\vtheta}(\vx_i)^\top\cE_{\vtheta}(\va_p) + \gamma]_+  \\  
\cR_z(\vtheta) &=  \textstyle
 \sum \limits_{l = 1}^L\sum\limits_{\substack{p: e_{lp} = 1\\n: e_{ln} = 0}} [\cE_{\vtheta}(\vz_l)^\top\cE_{\vtheta}(\va_n) - \cE_{\vtheta}(\vz_l)^\top\cE_{\vtheta}(\va_p) + \gamma]_+
\end{align*}
Here, $p$ is the positive anchor and $n$ are in-batch negatives anchors (explained in next section). Note that these two regularizers encourage the encoder to keep data points and labels closely embedded to their related anchor points and far away from unrelated anchor points. If we have more than one anchor set, say $\cA^1, \cA^2$, we can define corresponding regularizers $\cR^t_x(\vtheta), \cR^t_z(\vtheta), t{=} 1,2$. Note that, while we are explaining using two anchor sets, \alg can be easily extended to multiple anchor sets without the loss of generality.

\noindent\textbf{Training \alg:} \alg utilizes in-batch negative mining~\citep{Guo19,Dahiya21b,Dahiya23,Faghri18,Chen20,He20}(Figure~\ref{fig:training}). Specifically, a mini-batch is created using uniformly chosen set of queries and for each query, a relevant ad and a related anchor from query-anchor graph is chosen randomly. Similarly, for each of the chosen ad, a related anchor from ad-anchor graph is chosen randomly. Then, hard negative ads for a query are chosen only amongst those ads present in that mini-batch. Similarly, hard negative anchors for query/ad are chosen from only those anchors present in that mini-batch. Without the loss of generality training of \alg can be scaled to any number of anchor sets. \alg uses dual encoder models like DistilBERT~\citep{Sanh2019DistilBERTAD} and uses graph regularized training framework as described above to learn robust query/ad representation.

\noindent\textbf{Inference with \alg:} Inference of \alg is same as that of any dual encoder model. During inference, query/ads representation is computed using the trained encoder and top-$k$ relevant ads based on cosine similarity. Encoder trained through proposed graph regularization sees the gains in accuracy due to additional metadata graph without computational cost of GCNs.

\alg introduces a novel approach to incorporate graph metadata into XC, addressing the limitations of GCN-based methods. Unlike GCNs, which rely on computationally expensive graph induction, \alg employs a more efficient dual encoder architecture (\figurename~\ref{fig:compare}). This architecture, coupled with a graph-based regularization, empowers the model to learn robust representations of queries and ads (Figure~\ref{fig:training}). By focusing on relevant anchor points for regularization during training, \alg eliminates the need to induce graph at inference time, eliminating the impact of noise at inference time. Additionally, the proposed method demonstrates enhanced ability to handle new entities as graph-based regularization improves the representation of tail queries/ads. These combination of factors results in improved accuracy and efficiency, making \alg favorable for real-world recommendation tasks.
\balance
% experiments
%   data sets
%   evaluation methods
%   testbed, policies, hyperparams
%   main observation: better than NGAME and others
%   ablations, benefit points

\section{Experiments}
\label{sec:exp}

In this section, we compare \alg against deployed algorithms in Bing Ads, as well as established baselines on public benchmark datasets on the XML Repository~\citep{XMLRepo}. In particular, this paper benchmarks \alg on the LF-WikiSeeAlsoTitles-320K dataset where the recommendation task is similar to Bing Ads. The dataset is curated from Wiki dumps \href{https://dumps.wikimedia.org/enwiki/20220520/}{(link)}. The scenario involves recommending related articles. Articles under the ``See Also'' section were used as ground-truth labels. Internal hyperlinks and category links were used to create two sets of metadata graphs -- one using hyperlinked Wikipedia articles as anchors and the other, using Wikipedia categories as anchors. On the public dataset, all queries in the test set are unseen, making this a \textit{cold start} problem, while on the Bing data, user queries could either belong to the head/tail, or be unseen (\textit{cold start}). In online A/B tests, 
 \alg shows significant improvement in revenue, click through rates (CTR) and impression yield (IY). In addition to that, \alg observes significant reduction in bad match rate (BMR) with respect to brands (Brand BMR) and locations (Location BMR). This is crucial for advertisers who want to ensure that their ads are shown only for relevant queries, based on brand terms and location context. Please refer to Tab.~\ref{tab:data_stats} of the appendix for dataset statistics. For details on evaluation metrics please refer to appendix~\ref{app:metrics}.

\noindent \textbf{Implementation details}: For public datasets we initialize the encoder with a pre-trained DistilBERT and fine-tune it. During training, we prune the metadata graph using the fine-tuned encoder to eliminated noise in the training graph. Table~\ref{tab:hyperparams} in the appendix summarizes all hyper-parameters used for each dataset. We reiterate that, even though \alg uses a graph at training time, inference does not require any such information, making it highly suitable for long-tail queries and ads. We compare variants of \alg against baseline XC and dense retrieval approaches. In particular, we consider \alg(ANCE) and \alg(NGAME) where ANCE and NGAME where used as base XC model, on top of which we add \alg's regularization terms during training. All \alg variants and most baseline variants use the PyTorch~\citep{Paszke17} framework and were trained on $4$ Nvidia V100 GPUs. DEXML~\citep{Gupta2023} was trained on $16$ Nvidia A100 GPUs. Refer to Appendix~\ref{sec:impl} for additional details.

Using this section we answer the following questions:
\begin{itemize}
    \item What is the impact of graph regularization on accuracy and quality for real world applications?
    \item What is the computational cost of training \alg over GCN based approach (OAK)?
    \item What key design choices in \alg which lead to its success?
\end{itemize}

\noindent \textbf{Case-study for Sponsored Search}: Matching user queries with relevant advertiser. For example, for the query "cheap nike shoes", a valid ad is "nike sneakers" but "adidas shoes" or "nike shorts" are not.

We study the effectiveness of \alg in this application by comparing it against the state-of-the-art encoder in production by conducting A/B test on live search-engine traffic. The click logs were mined to gather graph metadata for \alg, including two types of signals:
\begin{enumerate}[leftmargin=*]
    \item Co-session queries/Ads: Queries that were asked in the same search session by multiple users as well as Ads that were clicked together in same session. 
    \item Similar queries/ Ads: Queries/Ads that are slight perturbations of each other but have the similar search intent.
\end{enumerate}

\alg was trained on a dataset containing 540M training documents and 360M labels mined as described above but over a longer period to conduct an A/B test on the search engine. \alg was found to increase the Impression-Yield (relevant ad impressions per user query) by 0.70\% and the CTR (clicks through rates) by 0.86\% when compared against a control ensemble containing state-of-the-art embedding-based, generative, GCN, and XC algorithms Table~\ref{fig:quantile}. Gains in \alg are primarily from rare query deciles, as shown in Table~\ref{fig:quantile}, which is the intended decile for \alg. In addition to gains in revenue metrics, \alg leads to significant reduction in bad match rates in terms of brands and location by -4.8\% and -7.2\% respectively (Table~\ref{fig:bing_bmr}). The A/B test was conducted on 160+ countries and Table~\ref{tab:bing} shows that \alg leads to gains in 1.25-1.44\% increase in revenue, 0.3-0.66\% increase in CTR and 0.28-0.9\% increase in IY.

\noindent \textbf{Results on benchmark datasets}: Table~\ref{tab:results-benchmark} compares \alg variants with graph and XC methods on standard XC metrics like Propensity scored precision and nDCG. \alg is 5 points more accurate over the best baseline numbers. In particular \alg is 2-3 points more accurate than traditional graph-based methods. Additionally, the \alg variants are 3-4 points more accurate over OAK~\citep{Mohan2024} \& PINA~\citep{Chien2023}, which use both XC and graph metadata.

\noindent \textbf{Analysis of gains and computation cost}: These experiments where conducted on public benchmark datasets for reproducibility and access to broader audience. Recall that the GCN (OAK) two-stage retrieval pipeline can be noisy. Table~\ref{tab:oracleLinker} demonstrates that, if we replace the first stage with the oracle linker (graph induction with zero error), the performance of these graph-based methods starts to outperform \alg variants. However, \emph{the oracle linker is never available for a novel test point}, and \alg variants achieved a similar performance in a fraction of the cost of training and prediction time as shown in Table~\ref{tab:train_pred_time}. In addition to reduction in inference time, \alg requires 3.5$\times$ less model parameters compared to OAK, as can be seen in Table~\ref{tab:train_pred_time}, while being 2 absolute points more accurate in Precision@1. Table~\ref{tab:pred_text} shows that \alg(ANCE) could make tail label predictions such as ``Crown group'' which was a 
 missing label ground truth in the training data.

\begin{table}
\centering
\caption{Query decile wise-comparison of \alg versus an ensemble of deployed Bing Ads algorithms (a.k.a. Control)}
\adjustbox{max width=\textwidth}{ \tabcolsep 2pt
    \begin{tabular}{l|ccc}
    \toprule
        \textbf{Decile} & \textbf{$\%\Delta$ Revenue} & \textbf{$\%\Delta$ CTR} & \textbf{$\%\Delta$ IY} \\ \midrule
        \textbf{HEAD} & 0.42\% & 0.22\% & 0.27\% \\
        \textbf{TORSO} & 0.41\% & 0.30\% & 0.19\% \\
        \textbf{TAIL} & 0.52\% & 0.34\% & 0.24\% \\ \midrule 
        \textbf{TOTAL} & 1.35\% & 0.86\% & 0.70\% \\ \bottomrule
    \end{tabular}
    }
    \label{fig:quantile}
\end{table}

\begin{table}
\centering
\caption{\alg leads to significant reduction in brand as well as location bad match rate in comparison to Bing Ads control over 160+ countries.}
\adjustbox{max width=\textwidth}{ \tabcolsep 2pt
    \begin{tabular}{c|c}
    \toprule
     $\%\Delta$ Brand BMR & $\%\Delta$ Location BMR \\
     \midrule
     -4.8 & -7.2 \\
     \bottomrule

    \end{tabular}
    }
    \label{fig:bing_bmr}
\end{table}

\begin{table}
    \centering
    \caption{Results of \alg in comparison to Bing Ads control in over 160+ countries grouped in 4 Clusters by geography}
    \adjustbox{max width=\textwidth}{
    \begin{tabular}{l|ccc}
    \toprule
        \textbf{Market Cluster} & \textbf{$\%\Delta$ Revenue} & \textbf{$\%\Delta$ CTR} & \textbf{$\%\Delta$ IY} \\ \midrule
        \textbf{MC-1} & 1.25 & 0.31 & 0.28 \\ 
        \textbf{MC-2} & 1.44 & 0.64 & 0.19 \\
        \textbf{MC-3} & 1.46 & 0.66 & 0.92 \\
        \textbf{MC-4} & 1.27 & 0.63 & 0.58 \\ \bottomrule
    \end{tabular}
    }
    \label{tab:bing}
\end{table}

\begin{table}
        \centering
        \caption{Results using Oracle Linker for GCN Vs \alg(ANCE) on LF-WikiSeeAlsoTitles-320K.}
        \label{tab:oracleLinker}
        \resizebox{\linewidth}{!}{
        \begin{tabular}{l|ccccc}
        \toprule
            \textbf{Method} & \textbf{P@1}&\textbf{P@5}&\textbf{N@5}&\textbf{PSP@1}&\textbf{PSP@5}\\
            \midrule
            \alg~(ANCE) & 35.2 &	18.4 &	35.2 &	29.0 &	34.5 \\
            OAK & 33.7 & 17.1 & 34.4 & 25.8 &  30.8 \\
            OAK + Oracle & 38.9 & 19.4 & 40.4 & 29.7 & 34.8 \\
            \bottomrule
        \end{tabular}
        }
\end{table}

\begin{table}
    \centering
    \caption{\alg(ANCE)'s computation relative to baselines on LF-WikiSeeAlsoTitles-320K.}
        \label{tab:train_pred_time}
        \resizebox{\linewidth}{!}{
        \begin{tabular}{l|cccc}
        \toprule
        \textbf{Method} & \textbf{Train Time} & \textbf{Pred Time} & \textbf{Model size} & \textbf{P@1} \\
        \midrule
        \alg~(ANCE) & 1$\times$ & 1$\times$ & 1$\times$ & 35.2 \\
        OAK & 1.5$\times$	& 2$\times$ & 3.5$\times$ &	33.7  \\
        ANCE & 0.9$\times$ & 1$\times$ & 1$\times$ & 30.8   \\
        DEXML & 2.1$\times$ & 1$\times$ & 1$\times$ & 29.9   \\
        \bottomrule
        \end{tabular}
        }
\end{table}

\begin{table}
    \centering
    \caption{Results comparing \alg against XC and Graph-based baselines on the short-text \textit{LF-WikiSeeAlsoTitles-320K} dataset. \alg variants is up to 15\% more accurate as compared to both text-based and graph-based baselines. Here PSP, P and N refers to propensity score precision, precision and nDCG respectively.}
\adjustbox{max width=0.95\linewidth}{  \tabcolsep 3pt
\begin{tabular}{l|ccccc}
\toprule
 &  \textbf{PSP@1}	&  \textbf{PSP@5}	&  \textbf{~P@1~} &	 \textbf{~P@5~} &	 \textbf{~N@5~} \\
 \midrule				
\alg(ANCE)	 & \textbf{28.9}	 & \textbf{34.5}	 & 35.2	 & 18.4	 & 36.5\\
\alg(NGAME)	 & 28.6	 & 34.4	 & \textbf{35.46}	 & \textbf{18.5}	 & \textbf{36.8}\\
\midrule
OAK	 & 25.8	 & 30.8	 & 33.7	 & 17.1	 & 34.3\\
GraphSage	 & 21.5	 & 23.5	 & 27.3	 & 12.9	 & 28.3\\
GraphFormer	 & 19.2	 & 22.7	 & 21.9	 & 11.8	 & 24.1\\
\midrule
ANCE	 & 25.1	 & 28.7	 & 30.7	 & 15.3	 & 31.4\\
NGAME	 & 24.4	 & 29.8	 & 32.6	 & 16.6	 & 33.2\\
DEXA	 & 24.4	 & 28.6	 & 31.7	 & 15.8	 & 32.2\\
DEXML	 & 22.8	 & 25.6	 & 29.9	 & 14.7	 & 30.7\\
DECAF	 & 16.7	 & 21.1	 & 25.1	 & 12.8	 & 25.9\\
Parabel	 & 9.2	 & 11.8	 & 17.6	 & 8.5	 & 17.4\\
CascadeXML	 & 12.6	 & 17.6	 & 23.4	 & 12.1	 & 23.4\\
XR-Transformer	 & 10.1	 & 12.7	 & 19.4	 & 8.9	 & 18.5\\
AttentionXML	 & 9.4	 & 11.7	 & 17.5	 & 8.5	 & 17.1\\
Bonsai	 & 10.6	 & 13.7	 & 19.3	 & 9.5	 & 19.2\\
SiameseXML	 & 26.8	 & 30.3	 & 31.9	 & 16.2	 & 32.5\\
ECLARE	 & 22.1	 & 26.2	 & 29.3	 & 15.1	 & 30.2\\
ELIAS	 & 13.4	 & 17.6	 & 23.3	 & 11.8	 & 23.6\\

\bottomrule
    \end{tabular}
    }  
    \label{tab:results-benchmark}
\end{table}

\begin{table}
    \centering
    \caption{A subjective comparison of predictions made by \alg, the leading text-based method NGAME, and the leading graph-based method GraphFormers on LF-WikiSeeAlsoTitles-320K. Labels that are a part of the ground truth are formatted in black color. Labels not a part of the ground truth are formatted in light gray color. Relevant labels that are missing from the ground truth are marked in bold black. \alg(ANCE) could make predict highly relevant labels such as ``Crown group'', which were missing from the ground truth as well as omitted by other methods.}
    \label{tab:pred_text}
    \resizebox{0.9\linewidth}{!}{
    \begin{tabular}{l|p{0.7\linewidth}}
    \toprule
    \textbf{Method} & \textbf{Prediction} \\
    \midrule
    \multicolumn{2}{c}{Document: Clade} \\
    \midrule
    \alg~(ANCE) & Cladistics, Phylogenetics, \textbf{Crown group}, Paraphyly, Polyphyly \\
    \midrule
    ANCE &  Cladistics, \wpred{Linnaean taxonomy}, Polyphyly, \wpred{Paragroup}, \wpred{Molecular phylogenetics} \\
    \midrule
    OAK & Phylogenetic nomenclature, \wpred{Molecular phylogenetics}, \wpred{Haplotype}, Cladistics, \wpred{Paragroup} \\
    \bottomrule
    \end{tabular}
    }
    
\end{table}

\noindent \textbf{Ablations on design choices}: To understand the impact of noisy edges in metadata, experiment \emph{``-- No Pruning''} disabled the trimming of noisy edges using cosine similarity filtering. A 4\% loss in P@1 was observed which underscores the necessity of pruning unhelpful edges during training. \alg(ANCE) uses multiple meta-data graphs for both document and label. To ascertain the contributions of the query-anchor and ads-anchor metadata graphs, the ablations \emph{``-- No Doc. Graph''} and \emph{``-- No Lbl. Graph''} were conducted. These experiments reveal that information from these graphs plays a significant role, as disabling either leads to a 1.5--2\% reduction in P@1. The information from these graphs can be incorporated in baseline methods like ANCE. To understand its impact, experiment \emph{``AugGT''} trains ANCE with augmented ground truth. The ground truth was expanded by using label propagation wherein a label and a training point are linked by an edge if the label shares a neighbor in the metadata graph of the said training point. \alg(ANCE) outperformed the \emph{``AugGT''} setup by 15\%. This suggests that while leveraging graph information for ground truth enhancement is convenient, it may not be as effective due to noisy edges.

\begin{table}
\centering
\caption{Ablations were done using ANCE as base algorithm (\alg(ANCE)) on LF-WikiSeeAlsoTitles-320K to understand the impact of design choices on the quality of encoder training. \alg(ANCE)'s design choices are seen to be optimal and offer 2.5--13\% improvement in the P@1 metric over alternate design choices.}
\adjustbox{max width=\textwidth}{ \tabcolsep 2pt
    \begin{tabular}{l|ccccc}
    \toprule
     RAMEN & \textbf{P@1} & \textbf{P@3} & \textbf{P@5}  & \textbf{N@3}  & \textbf{N@5}\\
    \midrule
    RAMEN~(ANCE) &  35.2 &	24.1 &	18.3 &	35.3 &	36.5\\ 
    \midrule
    $-$ No Pruning & 31.3 & 18.9 & 12.8 & 31.43 & 31.5 \\ 
    $-$ No Doc. Graph & 29.7 & 17.5 & 12.5 & 30.7 & 30.8 \\ 
    $-$ No Lbl. Graph & 34.1 & 22.7 & 14.6 & 32.0 & 34.1 \\ 
    AugGT & 15.6 & 8.9 & 6.5 & 15.7 & 16.3 \\ 
    \bottomrule

    \end{tabular}
    }
    \label{tab:ablation}
\end{table}

\section{Conclusion}

This paper presents \alg, a novel approach to incorporating graph metadata into extreme classification tasks. \alg addresses the limitations of GCN-based methods by leveraging a more efficient dual encoder architecture and a robust graph regularization technique. The experimental results demonstrate the effectiveness of \alg in enhancing accuracy and computational efficiency particularly for tail queries and ads.

By comparing \alg with state-of-the-art methods on public benchmarks and real-world datasets like Bing Ads, we have shown that it consistently outperforms existing approaches. The proposed graph regularization technique effectively mitigates the impact of noise in the graph data, leading to improved performance. In particular \alg can yield gains of up to 1-1.5\% in Revenue as well as 0.86\% and 0.70\% in CTR and Impression Yield, respectively over Bing Ads control. \alg leads to enhanced user experience as well as satisfaction by significantly reducing bad match rates in brand and location sensitive queries by 4.8\% and 7.2\% respectively. Additionally, \alg's efficient architecture enables it to scale well to large-scale datasets. Performance gains in \alg over OAK (GCN based method) also comes with 70\% reduced model parameters, 50\% reduced inference time.

Future research directions include exploring the application of \alg to other domains, investigating further enhancements to the model architecture, and developing more efficient training and inference techniques.

\clearpage

\bibliographystyle{iclr2025_conference}
\bibliography{ms}
\clearpage
\clearpage
\appendix
\onecolumn 
\begin{center}
\bfseries \Large \algtitle \\
\large (Appendix)
\end{center}
\section{Implementation details}
\label{sec:impl}

Links obtained on the metadata graph from raw data suffer from missing links in much the same way there are missing labels in the ground truth. To deal with this, \alg performs a random walk with restart on each anchor node. The random walk was performed for 400 hops with a restart probability of 0.8, thus ensuring that the walk did not wander too far from the starting node. This random walk could also introduce noisy edges, leading to poor model performance. To deal with such edges, in-batch pruning was performed and edges to only those anchors were retained which had a cosine similarity of $>0$ based on the embeddings given the encoder. To get the encoder, \alg initialize the encoder with a pre-trained DistilBERT and fine-tuned it for 10 epochs(warmup phase) using unpruned metadata graphs. Then the metadata graphs were pruned using the fine-tuned encoder. Encoder fine-tuning was then was continued for 5 epochs using the pruned graphs after which the graphs were re-pruned. These alternations of 5 epochs of encoder fine-tuning followed by re-pruning were repeated till convergence. The learning rate for each bandit was set to $0.01$. Table~\ref{tab:hyperparams} in supplementary material summarizes all hyper-parameters for each dataset. It is notable that even though \alg uses a graph at training time, inference does not require any such information, making it highly suitable for long-tail queries. 

\section{Data stats}
\begin{table*}[ht]
    \centering
    \caption{Dataset statistics summary for benchmark datasets used by \alg. Entries marked with $\ddagger$ were not disclosed because the dataset is proprietary.}
    \label{tab:data_stats}
    \adjustbox{max width=\textwidth}{\tabcolsep 2pt
    \begin{tabular}{ccccccccc}
    \hline
        \begin{tabular}{c}
             \textbf{\# Train Pts}  \\
             $N$
        \end{tabular}&
        \begin{tabular}{c}
             \textbf{\# Labels}  \\
             $L$
        \end{tabular}&
        \begin{tabular}{c}
             \textbf{\# Test Pts}  \\
             $N'$ 
        \end{tabular}&
        \begin{tabular}{c}
             \textbf{Avg. docs.}  \\
             \textbf{per label}
        \end{tabular}&
        \begin{tabular}{c}
             \textbf{Avg. labels}  \\
             \textbf{per doc.}
        \end{tabular}&
        \textbf{Graph Types}&
        \begin{tabular}{c}
             \textbf{\# Graph Nodes}  \\
             $G$
        \end{tabular}&
        
        \begin{tabular}{c}
             \textbf{Avg. node}\\ \textbf{neighbors}  \\
             \textbf{per doc.}
        \end{tabular}&
        \begin{tabular}{c}
             \textbf{Avg. node}\\ \textbf{neighbors}  \\
             \textbf{per label}
        \end{tabular}\\ \midrule
        \multicolumn{9}{c}{
        \textbf{LF-WikiSeeAlsoTitles-320K} 
        } \\
        \midrule
        693,082 & 312,330 & 177,515 & 4.67 & 2.11 &
        \begin{tabular}{c}
            Hyperlink  \\
            Category
        \end{tabular} &
        \begin{tabular}{c}
            2,458,399   \\
            656,086
        \end{tabular}&
        \begin{tabular}{c}
            38.87  \\
             4.74
        \end{tabular}&
        \begin{tabular}{c}
            7.71  \\
            4.82
        \end{tabular} \\
        % \hline
        % \multicolumn{9}{c}{
        % \textbf{LF-WikiTitles-500K} 
        % } \\
        % \midrule
        % 1,813,391 & 501,070 & 783,743 & 17.15 & 4.74 &
        % \begin{tabular}{c}
        %     Hyperlink  \\
        %     Category
        % \end{tabular} &
        % \begin{tabular}{c}
        %     2,148,579  \\
        %     766,929
        % \end{tabular} &
        % \begin{tabular}{c}
        %     16.46 \\
        %     2.35
        % \end{tabular} &
        % \begin{tabular}{c}
        %     8.53  \\
        %     4.21
        % \end{tabular} \\
        % \hline
        % \multicolumn{9}{c}{
        % \textbf{LF-AmazonTitles-1.3M} 
        % } \\
        % \midrule
        % 2,248,619 & 1,305,265 & 970,237 & 38.24 & 22.20 &
        % \begin{tabular}{c}
        %     related\_items \\
        %     category
        % \end{tabular} &
        % \begin{tabular}{c}
        %     916269  \\
        %     17981
        % \end{tabular} &
        % \begin{tabular}{c}
        %      1.98 \\
        %      3.35
        % \end{tabular} &
        % \begin{tabular}{c}
        %      3.95 \\
        %      583.04
        % \end{tabular} \\
        \hline
    \end{tabular}
    }
    
\end{table*}
\begin{table}
	\caption{ Hyper-parameter values for \alg on all datasets to enable reproducibility. \alg code will be released publicly. Most hyperparameters were set to their default values across all datasets. LR is learning rate. Multiple clusters were chosen to form a batch hence $B > C$. Clusters were refreshed after $5$ epochs. Cluster size $C$ was doubled after every 25 epochs. Margin $\gamma=0.3$ was used for contrastive loss. For training M2 number of positive samples and negative samples were kept at 2 and 12 respectively. A cell containing the symbol $\uparrow$ indicates that that cell contains the same hyperparameter value present in the cell directly above it.}
	\label{tab:hyperparams} \centering
	\adjustbox{max width=\linewidth}{
		\begin{tabular}{l|ccccc}
			\toprule
			\textbf{Dataset} & 
			\makecell{\textbf{Batch}\\\textbf{Size} $S$} & 
			\makecell{\textbf{Encoder}\\ epochs} & 
			\makecell{\textbf{Encoder LR}\\ $LR_1$} & 
			\makecell{\textbf{\bert seq.}\\\textbf{len} $L_{max}$} \\
			\midrule
            LF-WikiSeeAlsoTitles-330K & 1024 & 300 & 0.0002 & 32 \\ 
            \bottomrule
    	\end{tabular}
	}
\end{table}

\section{Evaluation metrics}
\label{app:metrics}
\begin{itemize}
    \item Impression Yield or $\text{IY} = \frac{\text{Relevant Ads impressed}}{\text{Total number of queries}} \times 100$
    \item Click through rate or $\text{CTR} = \frac{\text{Number of Clicks}}{\text{Number of Impressions}} \times 100$
\end{itemize}

For more information about Precision@K (P@K) and nDCG@K (N@K) and their propensity scored variants, please refer to \citep{XMLRepo}.

\end{document}